\documentclass[fleqn,10pt]{wlscirep}
\usepackage{subcaption}
\usepackage{hyperref} 
\usepackage{graphicx}
\usepackage{multirow}
\usepackage{pgfplots}
\usepackage{pdflscape}
\usepackage[clean]{revdiff}
\author[1,2,$\bigstar$]{Ivo M. Baltruschat}
\author[3]{Hannes Nickisch}
\author[3]{Michael Grass}
\author[1,2]{Tobias Knopp}
\author[3]{Axel~Saalbach}
\affil[1]{Section for Biomedical Imaging, University Medical Center Hamburg-Eppendorf, Hamburg, Germany}
\affil[2]{Institute for Biomedical Imaging, Hamburg University of Technology, Hamburg, Germany}
\affil[3]{Philips Research, Hamburg, Germany}
\affil[$\bigstar$]{ \href{mailto:i.baltruschat@uke.de}{i.baltruschat@uke.de}  }
\keywords{X-Ray, Deep Learning, Convolutional Neural Networks}
\begin{document}
	\title{Comparison of Deep Learning Approaches for Multi-Label Chest X-Ray Classification}
	\begin{abstract}
		The increased availability of \rnew{labeled} X-ray image archives (e.g. \rold{the }ChestX-ray14 dataset\rold{ from the NIH Clinical Center}) has triggered a growing interest in deep learning techniques. To provide better insight into the different approaches, and their applications to chest X-ray classification, we investigate a powerful network architecture in detail: the ResNet-50. Building on prior work in this domain, we consider transfer learning with and without fine-tuning as well as the training of a dedicated X-ray network from scratch. To leverage the high spatial \rchange{resolutions}{resolution} of X-ray data, we also include an extended ResNet-50 architecture, and a network integrating non-image data (patient age, gender and acquisition type) in the classification process. \rtag{R1.6}\rnew{In a concluding experiment, we also investigate multiple ResNet depths (i.e. ResNet-38 and ResNet-101).}
		
		In a systematic evaluation, using 5-fold re-sampling and a multi-label loss function, we \rchange{evaluate}{compare} the performance of the different approaches for pathology classification by ROC statistics and analyze differences between the classifiers using rank correlation.
		Overall, we observe a considerable spread in the achieved performance and conclude that the X-ray-specific \rchange{ResNet-50}{ResNet-38}, integrating non-image data yields the best overall results. \rtag{R1.4}\rchange{In addition, retrospective network analysis by activation mapping demonstrates that one pathology is identified due to non-anatomic image content.}{Furthermore, class activation maps are used to understand the classification process, and a detailed analysis of the impact of non-image features is provided.}
		
	\end{abstract}
	\flushbottom
	\maketitle
	\thispagestyle{empty}

	\begin{figure}[!ht]
		\vspace{-2em}
		\centering
		\includegraphics[width=1\textwidth]{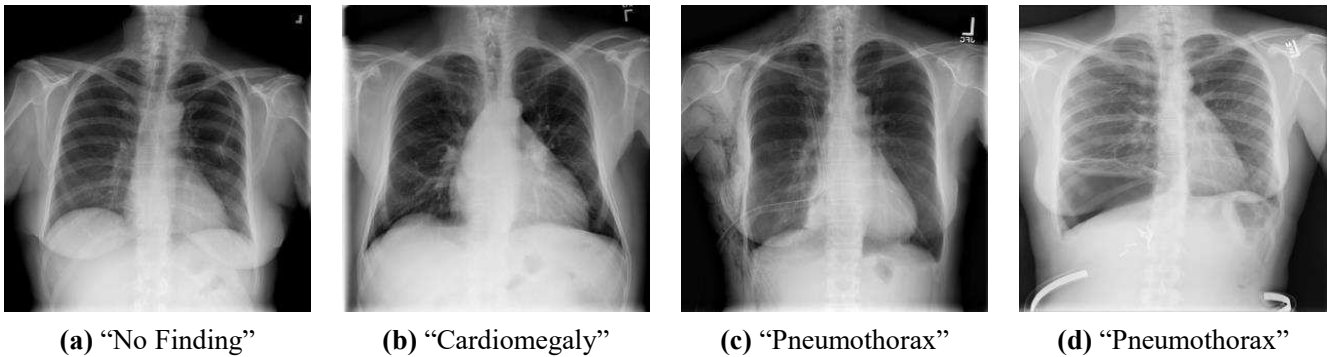}
		\caption[Four examples of the ChestX-ray14 dataset.]{Four examples of the ChestX-ray14 dataset. ChestX-ray14 consists of 112,120 frontal chest X-rays from 30,805 patients. All images are \rchange{labelled}{labeled} with up to 14 pathologies or ``No Finding''. The dataset does not only include acute findings, as the pneumothorax in figure (c), but also treated patients with a drain as ``pneumothorax'' (d).}
		\label{fig:cxr14}
		\vspace{-1em}
	\end{figure}
	\section{Introduction}
	\label{sec:intro}
	In the United Kingdom, the care quality commission recently reported that -- over the preceding 12 months -- a total of 23,000 chest X-rays (CXRs) were not formally reviewed by a radiologist or clinician at Queen Alexandra Hospital alone. Furthermore, three patients with lung cancer suffered significant harm because their CXRs had not been properly assessed \cite{CQC2017}. The Queen Alexandra Hospital is probably not the only hospital having problems with providing expert readings for every CXR. \rchange{Increasing}{Growing} populations and \rchange{life expectancies, is}{increasing life expectancies are} expected to drive an increase in demand for CXR readings. 
	
	In computer vision, deep learning has already shown its power for image classification with superhuman accuracy \cite{Krizhevsky2012,Szegedy2015,Simonyan2014,He2016}. In addition, the medical image processing field is vividly exploring deep learning. However, one major problem in the medical domain is the availability of large datasets with reliable ground-truth annotation. \rtag{R2.2}\rnew{Therefore, transfer learning approaches, as proposed by Bar et al.\cite{Bar2015}, were often considered to overcome such problems.}
	
	Two larger X-ray datasets have recently become available: the CXR dataset from Open-i \cite{Demner-Fushman2015} and \rchange{ChestX-ray14}{the ChestX-ray14 dataset} from the National Institutes of Health (NIH) Clinical Center \cite{Wang2017}. Figure~\ref{fig:cxr14} illustrates four selected examples from ChestX-ray14. Due to its size, the ChestX-ray14 consisting of 112,120 frontal CXR images from 30,805 unique patients attracted considerable attention in the deep learning community. Triggered by the work of Wang et al. \cite{Wang2017} using convolution neural networks (CNNs) from the computer vision domain, several research groups have begun to address the application of CNNs for CXR classification. In the work of Yao et al.\cite{Yao2017}, they presented a combination of a CNN and a recurrent neural network to exploit label dependencies. As a CNN backbone, they used a DenseNet \cite{Huang2017} model which was adapted and trained entirely on X-ray data. Li et al. \cite{Li2018} presented a framework for pathology classification and localization using CNNs. More recently, Rajpurkar et al. \cite{Rajpurkar2017} proposed transfer-learning with fine tuning, using a DenseNet-121 \cite{Huang2017}\rchange{ and}{, which raised} the AUC results on ChestX-ray14 for multi-label classification even higher. 
	
	\rchange{Unfortunately}{Unfortunately, a faithful} comparison of approaches remains difficult. Most reported results were obtained with differing experimental setups. This includes (among others) the employed network architecture, loss function and data augmentation. In addition, differing dataset splits were used and only Li et al. \cite{Li2018} reported 5-fold cross-validated results. In contrast to these results, our experiments (Sec.~\ref{sec:empirical}) demonstrate that performance of a network depends significantly on the selected split. \rtag{R1.5}\rtag{R2.5}\rtag{R2.7}\rnew{To have a fair comparison, Wang et al. \cite{Wang2017} released an official split later. Yao et al.\cite{yao2018weakly} and Guendel et al.\cite{guendel2018learning} reported results for this official split. While Guendel et al.\cite{guendel2018learning} hold the state-of-the-art results in all fourteen classes with a location-aware DenseNet-121.}
	
	\rchange{Henceforth, t}{T}o provide better insights into the effects of distinct design decisions for deep learning, we perform a systematic evaluation using a 5-fold re-sampling scheme. We empirically analyze three major topics:
	\begin{enumerate}
		\item weight initialization, pre-training and transfer learning (Section~\ref{sec:weight})
		\item network architectures such as ResNet-50 with large input size (Section~\ref{sec:archi})
		\item non-image features such as age, gender, and view position (Section~\ref{sec:meta})
	\end{enumerate}
	Prior work on ChestX-ray14 has been limited to the analysis of image data. In clinical practice however, radiologists employ a broad range of additional features during the diagnosis. To leverage the complete information of the dataset (i.e. age, gender, and view position), we propose in Section~\ref{sec:meta} a novel architecture integrating this information in addition to the learned image representation.

	\section{Methods}
	\label{sec:methods}
	In the following, we cast pathology detection as a multi-label classification problem. All images $X = \{\vec{x}_1, \dots, \vec{x}_N\}, \vec{x}_i \in \mathcal{X}$ are associated with a ground truth label $\vec{y}_i$, while we seek a classification function $\vec{f}: \mathcal{X} \rightarrow \mathcal{Y}$ that minimizes a specific loss function $\ell$ using $N$ training sample-label pairs $(\vec{x}_i,\vec{y}_i), i=1 \dots N$. Here, we encode the label for each image as a binary vector $\vec{y} \in \{0,1\}^M = \mathcal{Y}$  (with $M$ labels). We encode ``No Finding'' as an explicit additional label and hence have $M = 15$ labels. After an initial investigation of weighting loss functions such as positive/negative balancing \cite{Wang2017} and class balancing, we noticed no significant difference and decided to employ the class-averaged binary cross entropy (BCE) as our objective:
	\begin{equation}
	\ell (\vec{y}, \vec{f})= \frac{1}{M}\sum_{m=1}^{M} H[y_{m}, f_{m}], \; \text{with} \; H[y,f] = -y \log f -(1-y)\log (1-f).
	\end{equation}
	
	Prior work on the ChestX-ray14 dataset \rchange{focused}{concentrates} primarily on ResNet-50 and DenseNet-121 architectures. Due to its outstanding performance in the computer vision domain \cite{Huang2017}, we focus in our experiments on the ResNet-50 architecture \cite{He2016a}. To adapt the network to the new task, we replace the last dense layer of the original architecture with a new dense layer matching the number of labels and add a sigmoid activation function \rtag{R1.1}\rtag{R1.3}\rnew{for our multi-label problem (see Table~\ref{tab:ResNetArchitecture}).}
	
	\subsection{Weight Initialization and Transfer Learning}
	\label{sec:weight}
	We investigate two distinct initialization strategies for the ResNet-50. First, we follow the scheme described \rchange{in}{by He et al.} \cite{He2016}, where the network parameters are initialized with random values and thus the model is trained from scratch. Second, we initialize the network with pre-trained weights, where knowledge is \rchange{gained in}{transferred from} a different domain and task. Furthermore, we distinguish between \textit{off-the-shelf} (OTS) and \textit{fine-tuning} (FT) in the transfer-learning approach.
	
	A major drawback in medical image processing with deep learning is the limited size of datasets compared to the computer vision domain. Hence, training a CNN from scratch is often not feasible. One solution is transfer-learning. Following the notation in the work of Pan et al. \cite{Pan2010}, a source domain $\mathcal{D}_s = \{\mathcal{X}_s, P_s(X_s)\}$ with task $\mathcal{T}_s = \{\mathcal{Y}_s, f_s(\cdot)\}$ and a target domain $\mathcal{D}_t = \{\mathcal{X}_t, P_t(X_t)\}$ with task $\mathcal{T}_t = \{\mathcal{Y}_t, f_t(\cdot)\}$ are given with $\mathcal{D}_s\neq\mathcal{D}_t$ and/or $\mathcal{T}_s \neq \mathcal{T}_t$. 
	In transfer-learning, the knowledge gained in $\mathcal{D}_s$ and $\mathcal{T}_s$ is used to help learning \rchange{of the}{a} prediction function $f_t(\cdot)$ in $\mathcal{D}_t$. 
	
	Employing an off-the-shelf approach \cite{Yosinski2014, SharifRazavian2014}, the pre-trained network is used as a feature extractor, and only the weights of the last (classifier) layer are adapted. In fine-tuning, one chooses to re-train one or more layers with samples from the new domain. For both approaches, we use the weights of a ResNet-50 network trained on ImageNet as a starting point \cite{Russakovsky2015}.
	\rtag{R1.1}\rtag{R1.3}\rnew{In our fine-tuning experiment, we retrained all conv-layers as shown in Table~\ref{tab:ResNetArchitecture}.}
	\rmove{
	\begin{table}[!ht]
		\caption[Architecture of the original, off-the-shelf, and fine-tuned ResNets.]{\label{tab:ResNetArchitecture}\rtag{R1.1}\rtag{R1.3}\rnew{Architecture of the original, off-the-shelf, and fine-tuned ResNet-50. In our experiments, we use the ResNet-50 architecture and this table shows differences between the original architecture and ours (off-the-shelf and fine-tuned ResNet-50). If there is no difference to the original network, the word ``same'' is written in the table. The violet and bold text emphasizes, which parts of the network are changed for our application. All layers do employ automatic padding (i.e. depending on the kernel size) to keep spatial size the same. The conv3\_0, conv4\_0, and conv5\_0 layers perform a down-sampling of the spatial size with a stride of 2.}}
		\vspace{1ex}
		\centering
		\begin{footnotesize}
			\begin{tabular}{c |c | c | c | c }
				\textbf{Layer name}  & \textbf{Output size} & \textbf{Original 50-layer} & \textbf{Off-the-shelf} & \textbf{Fine-tuned} \\
				\hline \hline
				conv1 	& $112 \times 112$ & $7 \times 7$, 64-d, stride 2 & same & \color{violet}\textbf{fine-tuned}\\
				\hline
				pooling1 	& $56 \times 56$ & $3 \times 3$, 64-d, max pool, stride 2  & same & same\\
				\hline
				conv2\_x & $56 \times 56$   & $\begin{bmatrix} 1 \times 1 , \text{ 64-d, stride 1} \\ 3 \times 3 , \text{ 64-d, stride 1} \\ 1 \times 1, \text{ 256-d, stride 1}  \end{bmatrix} \times 3 $  & same & \color{violet}\textbf{fine-tuned}\\
				\hline
				conv3\_0 & $28 \times 28$   & $\begin{bmatrix} 1 \times 1, \text{ 128-d, stride 2} \\ 3 \times 3, \text{ 128-d, stride 1} \\ 1 \times 1, \text{ 512-d, stride 1}  \end{bmatrix} $  & same & \color{violet}\textbf{fine-tuned} \\
				\hline
				conv3\_x & $28 \times 28$   & $\begin{bmatrix} 1 \times 1, \text{ 128-d, stride 1} \\ 3 \times 3, \text{ 128-d, stride 1} \\ 1 \times 1, \text{ 512-d, stride 1}  \end{bmatrix} \times 3 $  & same & \color{violet}\textbf{fine-tuned} \\
				\hline
				conv4\_0 & $14 \times 14$   & $\begin{bmatrix} 1 \times 1, \text{ 256-d, stride 2} \\ 3 \times 3, \text{ 256-d, stride 1} \\ 1 \times 1, \text{ 1024-d, stride 1} \end{bmatrix} $  & same & \color{violet}\textbf{fine-tuned}\\
				\hline
				conv4\_x & $14 \times 14$   & $\begin{bmatrix} 1 \times 1, \text{ 256-d, stride 1} \\ 3 \times 3, \text{ 256-d, stride 1} \\ 1 \times 1, \text{ 1024-d, stride 1} \end{bmatrix} \times 5 $  & same & \color{violet}\textbf{fine-tuned}\\
				\hline
				conv5\_0 & $7 \times 7$     & $\begin{bmatrix} 1 \times 1, \text{ 512-d, stride 2} \\ 3 \times 3, \text{ 512-d, stride 1} \\ 1 \times 1, \text{ 2048-d, stride 1} \end{bmatrix}$  & same & \color{violet}\textbf{fine-tuned}\\
				\hline
				conv5\_x & $7 \times 7$     & $\begin{bmatrix} 1 \times 1, \text{ 512-d, stride 1} \\ 3 \times 3, \text{ 512-d, stride 1} \\ 1 \times 1, \text{ 2048-d, stride 1} \end{bmatrix} \times 2 $  & same & \color{violet}\textbf{fine-tuned}\\
				\hline
				pooling2 & $1 \times 1$ & $7 \times 7$, 2048-d, average pool, stride 1  & same & same\\
				\hline
				dense & $1 \times 1$  & 1000-d, dense-layer & \multicolumn{2}{c}{\color{violet}\textbf{15-d, dense-layer}} \\
				\hline
				loss & $1 \times 1$  & 1000-d, softmax & \multicolumn{2}{c}{\color{violet}\textbf{15-d, sigmoid, BCE}}\\
				\hline
			\end{tabular}
		\end{footnotesize}
		\vspace{2ex}
	\end{table}
}
	
	\subsection{Architectures}
	\label{sec:archi}
	In addition to the original ResNet-50 architecture, we employ two variants\rchange{.}{:} First, we reduce the number of input channels to one (the ResNet-50 is designed for the processing of RGB images from the ImageNet dataset), which should facilitate the training of an X-ray specific CNN. Second, we increase the input size by a factor of two (i.e. $448 \times 448$). To keep the model architectures similar, we only add a new \rchange{pooling}{max-pooling} layer after the first bottleneck block.\rtag{R1.1}\rtag{R1.3}\rnew{ This max-pooling layer has the same parameters as the ``pooling1'' layer (i.e. $3 \times 3$ kernel, stride 2, and padding). In Figure~\ref{fig:model}, our changes are illustrated at the image branch. A higher effective resolution could be beneficial for the detection of small structures, which could be indicative of a pathology (e.g. masses and nodules).}\rtag{R1.2}\rnew{ In the following, we use the postfix ``-1channel'' and ``-large'' to refer to our model changes.}
	
	\rtag{R1.1}\rtag{R1.3}\rtag{R1.6}\rnew{Finally, we investigate different model depths with the best performing setup. First, we implement a shallower ResNet-38 where we reduce the number of bottleneck blocks for conv2\_x, conv3\_x, and conv4\_x down to two, two, and three, respectively. Secondly, we also test the ResNet-101 and increased the number of conv\_3 blocks from 5 to 22 compare to the ResNet-50.}
	
	\subsection{Non-Image Features}
	\label{sec:meta}
	ChestX-ray14 contains information about the patient age, gender, and view position (i.e. if the X-ray image is acquired posterior-anterior (PA) or anterior-posterior (AP)). Radiologists use information beyond the image to conclude which pathologies are present or not. 
	The view position changes the expected position of organs in the X-ray images (i.e. PA images are horizontally flipped compared to AP). In addition, organs (e.g. the heart) are magnified in an AP projection as the distance to the detector is increased.
	
	As illustrated in Figure~\ref{fig:model}, we concatenate the image feature vector (i.e. output of the last pooling layer with dimension $2024 \times 1$) with the new non-image feature vector (with dimension $3 \times 1$). Therefore, view position and gender is encoded as $\{0, 1\}$ and the age is linearly scaled $[\text{min}({X}_{\text{pa}}), \text{max}({X}_{\text{pa}})] \mapsto [0, 1]$, in order to avoid a bias towards features with a large range of values.\rtag{R1.2}\rnew{ In our experiments, we used ``-meta'' to refer our model architecture with non-image features.}
	
	\rmove{\begin{figure}[t]
		
		\centering
		\includegraphics[width=\textwidth]{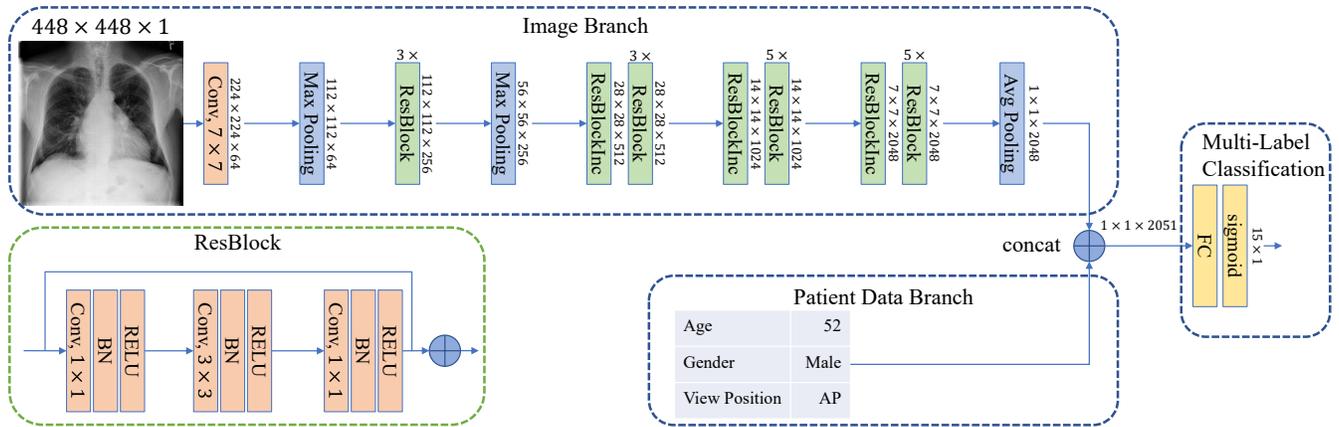}
		\caption{Patient-data adapted model architecture\rchange{.}{: ResNet-50-large-meta.} Our architecture is based on the ResNet-50 model. Because of the enlarged input size, we added a max-polling layer after the \rchange{first ResBlocks}{first three ResBlocks}. In addition, we fused image features and patient features at the end of our model to incorporate patient information.}
		\label{fig:model}
	\end{figure}}
	
	\subsection{ChestX-ray14 Dataset}
	\label{sec:data}
	\rtag{R2.4}\rnew{To evaluate our approaches for multi-label pathology classification, the entire corpus of ChestX-ray14 (Figure~\ref{fig:cxr14}) is employed. In total, the dataset contains 112,120 frontal chest X-rays from 30,805 patients. The dataset does not include the original DICOM images but Wang et al. \cite{Wang2017} performed a simple preprocessing based on the encoded display settings while the pixel depth was reduced to 8-bit. In addition, each image was resized to $1024\times1024$ pixel without preserving the aspect ratio.
	In Table~\ref{tab:distribution} and Figure~\ref{fig:age-dis}, we show the distribution of each class and the statistics for non-image information. The prevalence of individual pathologies are generally low and varies between $0.2\%$ and $17.74\%$ as shown in Table \ref{tab:disease-dis}. While, the distribution of patient gender and view position is quite even with a ratio of 1.3 and 1.5, respectively (see Table~\ref{tab:meta-dis}). In Figure~\ref{fig:age-dis}, the histogram shows the distribution of patient age in ChestX-ray14. The average patient age is $46.87$ years with a standard deviation of $16.60$ years.}
	
	\rtag{R1.4}\rtag{R2.1}\rnew{To determine if the provided non-image features contain information for a disease classification, we performed an initial experiment. We trained a very simple Multi-layer Perceptron (MLP) classifier only with the three non-image feature as input. The MLP classifier has a low average AUC of 0.61 but this still indicates that those non-image features could help to improve classification results when provided to our novel model architecture.}
	\rmove{\begin{table}
		\centering
		\setlength{\tabcolsep}{5pt}
		\caption[Overview of label distributions]{\rtag{R2.1}\rnew{Overview of label distributions in the ChestX-ray14 dataset.}}
		\label{tab:distribution}
		\begin{subtable}[t]{0.45\linewidth}
			\centering
			\caption[D]{\rtag{R2.1}\rnew{Diseases}}
			\label{tab:disease-dis}
			\begin{tabular}{l r r r }
				\noalign{\smallskip}
				\textbf{Pathology} & True & False & Prevalence $[\%]$ \\
				\noalign{\smallskip}
				\hline
				\noalign{\smallskip}	
				\textbf{Cardiomegaly}
				& $2,776$
				& $109,344$
				& $2.48$\\
				
				\textbf{Emphysema}
				& $2,516$
				& $109,604$
				& $2.24$\\
				
				\textbf{Edema}
				& $2,303$
				& $109,817$
				& $2.05$\\
				
				\textbf{Hernia}
				& $227$
				& $111,893$
				& $0.20$\\
				
				\textbf{Pneumothorax}
				& $5,302$
				& $106,818$
				& $4.73$\\
				
				\textbf{Effusion}
				& $13,317$
				& $98,803$
				& $11.88$\\
				
				\textbf{Mass}
				& $5,782$
				& $106,338$
				& $5.16$\\
				
				\textbf{Fibrosis}
				& $1,686$
				& $110,434$
				& $1.50$\\
				
				\textbf{Atelectasis}
				& $11,559$
				& $100,561$
				& $10.31$\\
				
				\textbf{Consolidation}
				& $4,667$
				& $107,453$
				& $4.16$\\
				
				\textbf{Pleural Thicken.}
				& $3,385$
				& $108,735$
				& $3.02$\\
				
				\textbf{Nodule}
				& $6,331$
				& $105,789$
				& $5.65$\\
				
				\textbf{Pneumonia}
				& $1,431$
				& $110,689$
				& $1.28$\\
				
				\textbf{Infiltration}
				& $19,894$
				& $92,226$
				& $17.74$
				
			\end{tabular}
		\end{subtable}
		\begin{subtable}[t]{0.45\linewidth}
			\centering
			\caption[MI]{\rtag{R2.1}\rnew{Meta-information}}
			\label{tab:meta-dis}
			\begin{tabular}{l r r r }
				\noalign{\smallskip}
				& Female & Male & Ratio \\
				\noalign{\smallskip}
				\hline
				\noalign{\smallskip}	
				\textbf{Patient Gender}
				& $63,340$
				& $48,780$
				& $1.30$\\
				\noalign{\bigskip}	
				& PA & AP & Ratio \\
				\noalign{\smallskip}
				\hline
				\noalign{\smallskip}	
				\textbf{View Position}
				& $67,310$
				& $44,810$
				& $1.50$
			\end{tabular}
		\end{subtable}
	
	\end{table}
	\begin{figure}
		\centering
		\includegraphics[width=\textwidth]{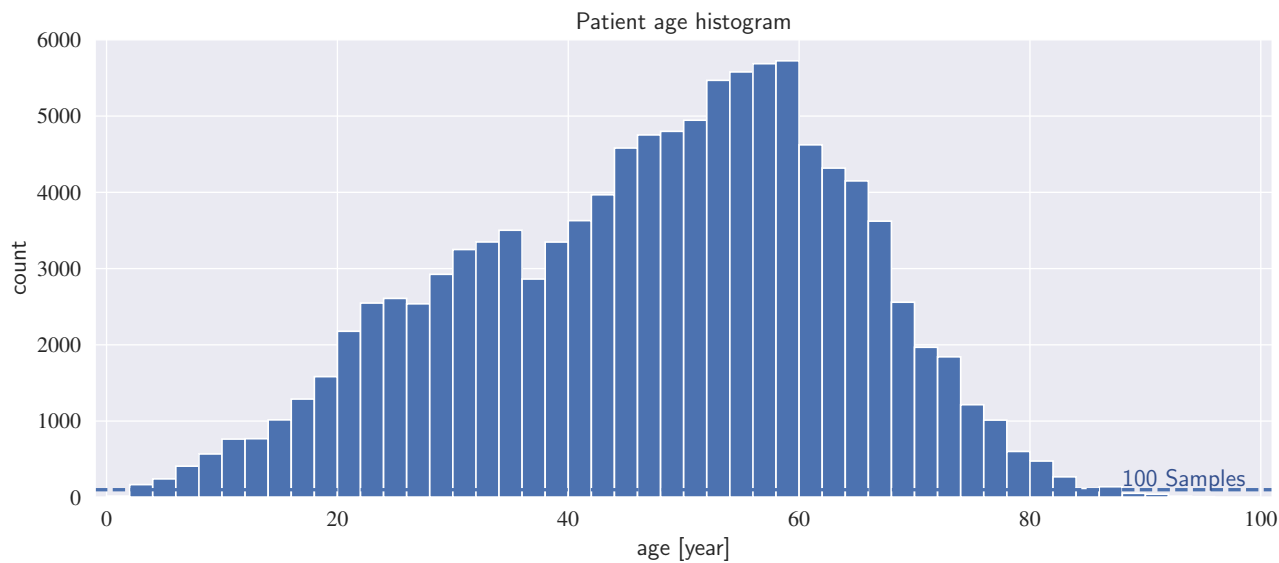}
		\caption[Distribution of patient age]{\rtag{R2.1}\rnew{Distribution of patient age in the ChestX-ray14 dataset. Each bin covers a width of two years. The average patient age is $46.87$ years with a standard deviation of $16.60$ years.}}
		\label{fig:age-dis}
		
	\end{figure}}

	\section{Experiments and Results}
	\label{sec:empirical}
	\rold{To evaluate our approaches for multi-label pathology classification, the entire corpus of ChestX-ray14 (Figure~\ref{fig:cxr14}) is employed. The dataset does not include the original DICOM images but Wang et al. \cite{Wang2017} performed a simple preprocessing where the intensity range was rescaled from a higher bit-depth down to 8-bit. In addition, each image was resized to $1024\times1024$ pixel without preserving the aspect ratio.}\rcomment{Moved to Section~\ref{sec:data}}
	For an assessment of the generalization performance, we perform a 5 times re-sampling scheme \cite{Molinaro2005}. Within each split, the data is divided into 70\% training, 10\% validation, and 20 \% testing. When working with deep learning, hyper-parameters, and tuning without a validation set and/or cross-validation can easily result in over-fitting. Since individual patients have multiple follow-up acquisitions, all data from a patient is assigned to a single subset only. This leads to a large patient number diversity (e.g. split two has 5,817 patients and 22,420 images whereas split 5 has 6,245 patients and the same number of images). We estimate the average validation loss over all re-samples to determine the best models. Finally, our results are calculated for each fold on the test set and averaged afterwards.
	
	\rtag{R1.5-6}\rtag{R2.5-7}\rnew{To have a fair comparison to other groups, we conduct an additional evaluation using the best performing architecture with different depth on the official split of Wang et al.\cite{Wang2017} in Section~\ref{sec:compare}.}
	
	\noindent \textbf{Implementation:} In all experiments, we use a fixed setup. To extend ChestX-ray14, we use the same geometric data augmentation as in the work of Szegedy et al. \cite{Szegedy2015}. At training, we sample various sized patches of the image with sizes between $8\%$ and $100\%$ of the image area. The aspect ratio is distributed evenly between $3:4$ and $4:3$. In addition, we employ random \rchange{rotation}{rotations} between $\pm7^{\circ}$ and horizontal flipping. For validation and testing, we rescale images to $256 \times 256$ and $480 \times 480$\rchange{ }{ pixels} for small and large spatial size, respectively. Afterwards, we use the center crop as input image. As in the work of He et al. \cite{He2016}, dropout is not employed \cite{Srivastava2014}. As optimizer, we use ADAM \cite{Kingma2015} with default parameters for $\beta_1 = 0.9$ and $\beta_2 = 0.999$. The learning rate $lr$ is set to $lr = 0.001$ and $lr = 0.01$ for transfer-learning and from scratch, respectively. While training, we reduce the learning rate by a factor of 2 when the validation loss does not improve. Due to model architecture variations, we use batch sizes of 16 and 8 for transfer-learning and from scratch with a large input size, respectively. The models are implemented in CNTK and trained on GTX 1080 GPUs yielding a processing time of around $10 \text{ms}$ per image.
	
	\noindent \textbf{Results:} Table~\ref{tab:compare} summarizes the outcome of our evaluation \rold{and we show state-of-the-art reference results in Figure~\ref{fig:compare}}. In total, we evaluate eight different experimental setups with varying weight initialization schemes and network architectures as well as with and without non-image features. We perform \rchange{a}{an} ROC analysis using the area under the curve (AUC) for all pathologies, compare the classifier scores by Spearman's pairwise rank correlation coefficient, and \rchange{employed}{employ} the state-of-the-art method Gradient-weighted Class Activation Mapping (Grad-CAM) \cite{Selvaraju2017} to \rchange{get}{gain} more insight into our CNNs. Grad-CAM is a method for visual\rchange{ explanations for predictions of CNN models}{ly assessing CNN model predictions}. The method highlights important regions in the input image for a specific classification \rchange{results}{result} by using the gradient of the final convolutional layer.

	The results indicate a high variability of the outcome with respect to the selected dataset split. Especially for ``Hernia'', which is the class with the smallest number of positive samples, we observe a standard deviation of up to 0.05. As a result, an assessment of existing approaches and \rchange{comparing}{comparison of} their performance is difficult, since prior work \rchange{focused}{focuses} mostly on a single (random) split.
	
	With respect to the different initialization \rchange{schemes}{schemes,} we observe already reasonable results for OTS networks that are optimized on natural images. Using fine-tuning\rold{ techniques}, the results are improved considerably, from $0.730$ to $0.819$ AUC on average. A complete training of the \rtchange{R1.2}{ResNet-50}{ResNet-50-1channel} using CXRs results in a rather comparable performance. Only the high-resolution variant of the \rtchange{R1.2}{ResNet-50}{ResNet-50-large} outperforms the FT approach by $0.002$ on average AUC. In particular, for smaller pathologies like \rchange{masses and nodules}{nodules and masses} an improvement is observed (i.e. \rchange{0.017}{0.018} and 0.006 AUC increase, respectively), while for other pathologies a similar, or slightly lower performance is estimated.
	
	Finally, all our experiments with non-image features slightly increase the AUC on average to its counterpart (i.e. without non-image feature). Our from scratch trained \rtchange{R1.2}{ResNet-50 with an enlarged input size and integrated non-image}{ResNet-50-large-meta} yields the best overall performance with $0.822$ average AUC.
	
	\rtag{R1.4}\rtag{R2.1}\rtag{R2.4}\rnew{To get a better insight why the non-image features only slightly increased the AUC for our fine-tuned and from scratch trained models, we investigated the capability of predict the non-image features based on the extracted image features. We used our from scratch trained model (i.e. ResNet-50-large) as a feature extractor and trained three models to predict the patient age, patient gender, and view position (VP) -- i.e. ResNet-50-large-age, ResNet-50-large-gender, ResNet-50-large-VP. We employed the same training setup as in our experiments before. First, our ResNet-50-large-VP model can predict with a very high AUC of $0.9983 \pm 0.0002$ the correct VP (i.e. we encoded AP as true and PA as false). After choosing the optimal threshold based on Youden index, we calculated a sensitivity and specificity of $99.3\%$ and $99.1\%$, respectively. Secondly, the ResNet-50-large-gender predicts the patient gender also very precisely with a high AUC of $0.9435 \pm 0.0067$. The sensitivity and specificity with $87.8\%$ and $85.9\%$ is also high. Finally, to evaluate the performance of the ResNet-50-large-age we report the mean absolute error (MAE) with standard deviation because age prediction is a regression task. The model achieved a mean absolute error of $9.13 \pm 7.05$ years. The results show that the image features already encode information about the non-image features. This might be the reason that our proposed model architecture with the non-image features at hand did not increased the performance by a large margin.}
	
	Furthermore, the similarity between the trained models in terms of their predictions was investigated. Therefore, Spearman?s rank correlation coefficient was computed for the predictions of all model pairs, and averaged over the folds. The pairwise correlations coefficients for the models are given in Table~\ref{tab:rankCorr}. Based on the degree of correlation, three groups can be identified. First, we note that the ``from scratch models'' (i.e. ``1channel'' and ``large'') without non-image features have the highest correlation of 0.93 amongst each other, followed by the fine-tuned models with 0.81 and 0.80 for ``1channel'' and ``large'', respectively. Second, the OTS model surprisingly has higher correlation with the from scratch models than the fine-tuned model. Third, for models with non-image feature, no such correlation is observed and their value is between 0.32 to 0.47. This indicates that models which have been trained exclusively on X-ray data achieve not only the highest accuracy, but are furthermore most consistent.
	\begin{table}[!ht]
		\setlength{\tabcolsep}{5pt}
		\centering
		\caption{AUC result overview for all our experiments. In this table, we present averaged results over all five splits and the calculated standard deviation (std) for each pathology. We divide our experiments into three categories. First, without and with non-image features. Second, transfer-learning with off-the-shelf (OTS) and fine-tuned (FT) models. Third, from scratch where ``1channel'' refers to same input size as in transfer-learning but changed number of channels. ``large'' means we changed the input dimensions to $448 \times 448 \times 1$. For better comparison, we present the average AUC and the standard deviation over all pathologies in the last row. Bold text emphasizes the overall highest AUC value. Values are scaled by 100 for convenience.}
		\label{tab:compare}
		\begin{tabular}{l c c c c | c c c c }
			\noalign{\smallskip}
			& \multicolumn{4}{c|}{Without non-image features} & \multicolumn{4}{c}{With non-image features} \\
			\textbf{Pathology} & OTS & FT & 1channel & large & OTS & FT & 1channel & large \\
			\noalign{\smallskip}
			\hline
			\noalign{\smallskip}	
			\textbf{Cardiomegaly}
			& $72.7 \pm 1.8$
			& $88.5 \pm 0.7$
			& $88.9 \pm 0.5$
			& $89.7 \pm 0.3$
			& $75.9 \pm 1.4$
			& $88.4 \pm 0.8$
			& $\mathbf{90.2 \pm 0.4}$
			& $89.8 \pm 0.8$ \\
			
			\textbf{Emphysema}
			& $77.8 \pm 2.1$
			& $89.2 \pm 1.0$
			& $87.0 \pm 0.8$
			& $88.3 \pm 1.3$
			& $79.8 \pm 1.9$
			& $\mathbf{89.4 \pm 1.2}$
			& $87.4 \pm 1.3$
			& $89.1 \pm 1.2$ \\
			
			\textbf{Edema}
			& $84.4 \pm 0.6$
			& $\mathbf{89.1 \pm 0.4}$
			& $\mathbf{89.1 \pm 0.6}$
			& $88.8 \pm 0.5$
			& $85.7 \pm 0.5$
			& $\mathbf{89.1 \pm 0.7}$
			& $89.0 \pm 0.6$
			& $88.9 \pm 0.3$ \\
			
			\textbf{Hernia}
			& $78.8 \pm 1.4$
			& $85.5 \pm 3.8$
			& $88.1 \pm 4.2$
			& $87.5 \pm 4.5$
			& $81.9 \pm 2.5$
			& $88.2 \pm 3.2$
			& $89.3 \pm 4.4$
			& $\mathbf{89.6 \pm 4.4}$ \\
			
			\textbf{Pneumothorax}
			& $77.3 \pm 1.3$
			& $\mathbf{87.0 \pm 0.8}$
			& $85.7 \pm 0.9$
			& $85.9 \pm 0.9$
			& $79.1 \pm 1.2$
			& $86.5 \pm 0.6$
			& $85.4 \pm 0.7$
			& $85.9 \pm 1.1$ \\
			
			\textbf{Effusion}
			& $79.4 \pm 0.4$
			& $87.1 \pm 0.2$
			& $\mathbf{87.6 \pm 0.2}$
			& $\mathbf{87.6 \pm 0.2}$
			& $80.6 \pm 0.4$
			& $87.2 \pm 0.3$
			& $\mathbf{87.6 \pm 0.2}$
			& $87.3 \pm 0.3$ \\
			
			\textbf{Mass}
			& $66.8 \pm 0.6$
			& $82.2 \pm 1.0$
			& $83.3 \pm 0.6$
			& $\mathbf{83.9 \pm 0.9}$
			& $68.6 \pm 0.6$
			& $82.2 \pm 1.0$
			& $83.3 \pm 0.7$
			& $83.2 \pm 0.3$ \\
			
			\textbf{Fibrosis}
			& $72.0 \pm 0.9$
			& $\mathbf{80.0 \pm 0.9}$
			& $79.9 \pm 0.8$
			& $79.2 \pm 1.6$
			& $73.9 \pm 0.8$
			& $\mathbf{80.0 \pm 0.9}$
			& $79.6 \pm 0.5$
			& $78.9 \pm 0.5$ \\
			
			\textbf{Atelectasis}
			& $71.8 \pm 0.6$
			& $\mathbf{80.3 \pm 0.7}$
			& $79.9 \pm 0.4$
			& $79.2 \pm 0.7$
			& $73.2 \pm 0.7$
			& $80.1 \pm 0.6$
			& $79.3 \pm 0.6$
			& $79.1 \pm 0.4$ \\
			
			\textbf{Consolidation}
			& $74.3 \pm 0.3$
			& $79.5 \pm 0.5$
			& $\mathbf{80.6 \pm 0.4}$
			& $80.0 \pm 0.3$
			& $75.3 \pm 0.3$
			& $79.6 \pm 0.5$
			& $80.4 \pm 0.5$
			& $80.0 \pm 0.7$ \\
			
			\textbf{Pleural Thicken.}
			& $68.8 \pm 1.0$
			& $\mathbf{79.0 \pm 0.7}$
			& $78.4 \pm 0.9$
			& $78.0 \pm 1.1$
			& $70.8 \pm 1.1$
			& $78.6 \pm 1.1$
			& $78.2 \pm 1.3$
			& $77.1 \pm 1.3$ \\
			
			\textbf{Nodule}
			& $65.0 \pm 0.8$
			& $72.6 \pm 0.9$
			& $73.3 \pm 0.8$
			& $75.1 \pm 1.3$
			& $66.5 \pm 0.7$
			& $74.7 \pm 0.6$
			& $74.0 \pm 0.7$
			& $\mathbf{75.8 \pm 1.4}$ \\
			
			\textbf{Pneumonia}
			& $66.4 \pm 2.7$
			& $74.4 \pm 1.6$
			& $74.3 \pm 1.5$
			& $75.3 \pm 2.2$
			& $68.3 \pm 2.3$
			& $73.3 \pm 1.3$
			& $74.8 \pm 1.5$
			& $\mathbf{76.7 \pm 1.5}$ \\
			
			\textbf{Infiltration}
			& $65.9 \pm 0.2$
			& $69.9 \pm 0.6$
			& $\mathbf{70.2 \pm 0.3}$
			& $\mathbf{70.2 \pm 0.5}$
			& $67.0 \pm 0.4$
			& $\mathbf{70.2 \pm 0.2}$
			& $70.1 \pm 0.5$
			& $70.0 \pm 0.7$\\
			
			\hline
			\noalign{\smallskip}
			
			\textbf{Average}
			& $73.0 \pm 1.1$
			& $81.7 \pm 1.0$
			& $81.9 \pm 0.9$
			& $82.1 \pm 1.2$
			& $74.8 \pm 1.1$
			& $82.0 \pm 0.9$
			& $82.0 \pm 1.0$
			& $\mathbf{82.2 \pm 1.1}$\\
			
			\textbf{No Findings}
			& $71.6 \pm 0.3$
			& $76.9 \pm 0.5$
			& $\mathbf{77.3 \pm 0.3}$
			& $77.1 \pm 0.4$
			& $72.5 \pm 0.3$
			& $76.8 \pm 0.4$
			& $77.1 \pm 0.4$
			& $77.1 \pm 0.3$\\
			
		\end{tabular}
	\end{table}
	\begin{table}[!ht]
		\centering
		\caption{Spearman's rank correlation coefficient is calculated between all model pairs and is averaged over all five splits. Our experiments are grouped into three categories. First, ``Without'' and ``With'' non-image features. Second, transfer-learning with off-the-shelf (OTS) and fine-tuned (FT) models. Third, from scratch where ``1channel'' refers to same input size as in transfer-learning but changed number of channels. ``large'' means we changed the input dimensions to $448 \times 448 \times 1$. We identify three clusters: all models under ``With'', models trained from scratch and ``Without'', and the ``OTS'' model.}
		\label{tab:rankCorr}
		\begin{tabular}{l l c c c c c c c c }
			\noalign{\smallskip}
			& & \multicolumn{4}{c}{Without} & \multicolumn{4}{c}{With} \\
			& & OTS & FT & 1channel & large & OTS & FT & 1channel & large \\
			\noalign{\smallskip}
			\hline
			\noalign{\smallskip}
			\multirow{4}{*}{Without\;}&
			OTS            & -  & 0.65      & 0.74   & 0.73      & 0.46     & 0.38           & 0.40        & 0.59\\
			&FT             & 0.65  & -      & 0.81   & 0.80      & 0.38     & 0.42           & 0.43        & 0.64\\
			&1channel       & 0.74  & 0.81      & -   & 0.93      & 0.41     & 0.43           & 0.47        & 0.71\\
			&large          & 0.73  & 0.80      & 0.93   & -      & 0.40     & 0.43           & 0.47        & 0.71\\
			
			\multirow{4}{*}{With}&
			OTS            & 0.46  & 0.38      & 0.41   & 0.40      & -     & 0.32           & 0.33        & 0.39\\
			&FT             & 0.38  & 0.42      & 0.43   & 0.43      & 0.32     & -           & 0.35        & 0.42\\
			&1channel       & 0.40  & 0.43      & 0.47   & 0.47      & 0.33     & 0.35           & -        & 0.45\\
			&large          & 0.59  & 0.64      & 0.71   & 0.71      & 0.39     & 0.42           & 0.45        & -\\
			\hline
		\end{tabular}
	\end{table}
	
\begin{figure}[!ht]
		\centering

		\centering
		\includegraphics[width=0.8\textwidth]{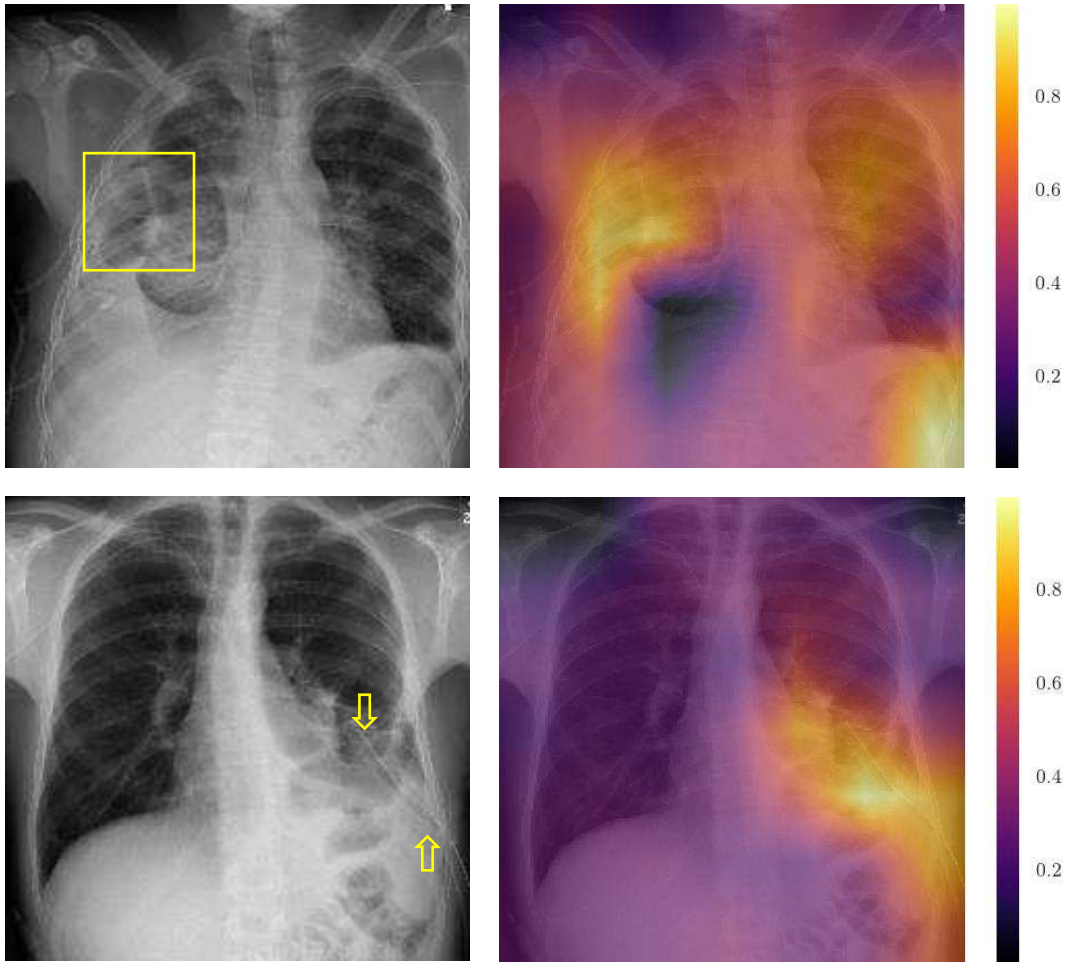}
		\caption{Grad-CAM result for two example images. In the first one, we marked the location of the pneumothorax with a yellow box. As shown in the Grad-CAM image next to it, the models highest activation for the prediction is within the correct area. The second row shows a negative example where the highest activation, which was responsible for the final predication ``pneumothorax'', is at the drain. This indicates that our trained CNN picked up drains as a main feature for ``pneumothorax''. We marked the drain with yellow arrows.}
		\label{fig:grad-cam}
	\end{figure}

	While our proposed network architecture achieves high AUC values in many categories of the ChestX-ray14 dataset, the applicability of such a technology in a clinical environment depends considerably on the availability of data for model training and evaluation. In particular for the NIH dataset the reported label noise\cite{Wang2017} and the medical interpretation of the label are an important issue. As mention by Luke Oakden-Rayner \cite{Luke2017}, the class ``pneumothorax'' is often labeled for already treated cases (i.e. a drain is visible in the image which is used to tread the pneumothorax) in the ChestX-ray14 dataset. We employ Grad-CAM to get an insight, if our trained CNN picked up the drain as a main feature for ``pneumothorax''. Grad-CAM visualizes the areas which are most responsible for the final prediction as a heatmap. In Figure~\ref{fig:grad-cam}, we show two examples of our testset where the highest activations are around the drain. This indicates that the network learned not only to detect an acute pneumothorax but also the presence of chest drains. Therefore, the utility of the ChestX-ray14 dataset for the development of clinical applications is still an open issue.
	
	\subsection{Comparison to other approaches}
	\label{sec:compare}
	\rtag{R1.5-6}\rtag{R2.5-7}\rnew{
	In our evaluation, we noticed a considerable spread of the results in terms of AUC values. Next to the employed data splits, this could be attributed to the (random) initialization of the models, and the stochastic nature of the optimization process.}
	
	\rnew{When ChestX-ray14 was made publicly available, only images and no official dataset splitting was released. Hence, researcher started to train and test their proposed methods on their own dataset split. We noticed a large diversity in performance with different splits of our re-sampling. Therefore, a direct comparison to other groups might be miss leading in the sense of state-of-the-art results. For example, Rajpurkar et al. \cite{Rajpurkar2017} reported state-of-the-art results for all 14 classes on their own split. In Figure~\ref{fig:compare}, we compare our best performing model architecture (i.e. ResNet-50-large-meta) of the re-sampling experiments to Rajpurkar et al. and other groups. For our model, we plot the minimum and maximum AUC over all re-samplings as error bars to illustrate the effect of random splitting. We achieve state-of-the-art results for ``effusion'' and ``consolidation'' when directly comparing our AUC (i.e. averaged over 5 times re-sampling) to former state-of-the-art results. Comparing the maximum AUC over all re-sampling splits results in state-of-the-art performance for ``effusion'', ``pneumonia'', ``consolidation'', ``edema'', and ``hernia'' and indicates that a fair comparison between groups without the same splitting might be non conclusive.}
	
	\rnew{Later, Wang et al.\cite{Wang2017} released an official split of the ChestX-ray14 dataset. To have a fair comparison to other groups, we report results on this split for our best performing architecture with different depths -- ResNet-38-large-meta, ResNet-50-large-meta, and ResNet-101-large-meta -- in Table~\ref{tab:compareOfficial}. 
	First, we compare our results to Wang et al.\cite{Wang2017} and Yao et al.\cite{yao2018weakly} because Guendel et al.\cite{guendel2018learning} used an additional dataset -- PLCO dataset\cite{gohagan2000prostate} -- with 185,000 images. While the ResNet-101-large-meta already has a higher average AUC with $0.785$ and in 12 out of 14 classes a higher individual AUC, the performance is compared to our ResNet-38-large-meta and ResNet-50-larg-meta lower. Reducing the number of layers increased the averaged AUC from $0.785$ to $0.795$ and $0.806$ for ResNet50-large-meta and ResNet38-larg-meta, respectively. Hence, our results indicate that training a model with less parameter on Chest-Xray14 is beneficial for the overall performance. 
	Secondly, Guendel et al.\cite{guendel2018learning} reported state-of-the-art results for the official split in all 14 classes with an averaged AUC of $0.807$. While our ResNet-38-large-meta is trained with 185,000 images less, it still achieved state-of-the-art results for ``Emphysema'', ``Edema'', ``Hernia'', ``Consolidation'', and ``Pleural Thicken.'' and a slight less average AUC of $0.806$.}
\rmove{
	\begin{figure}
		\vspace{-1.5em}
		\centering
		\includegraphics[width=0.7\textwidth]{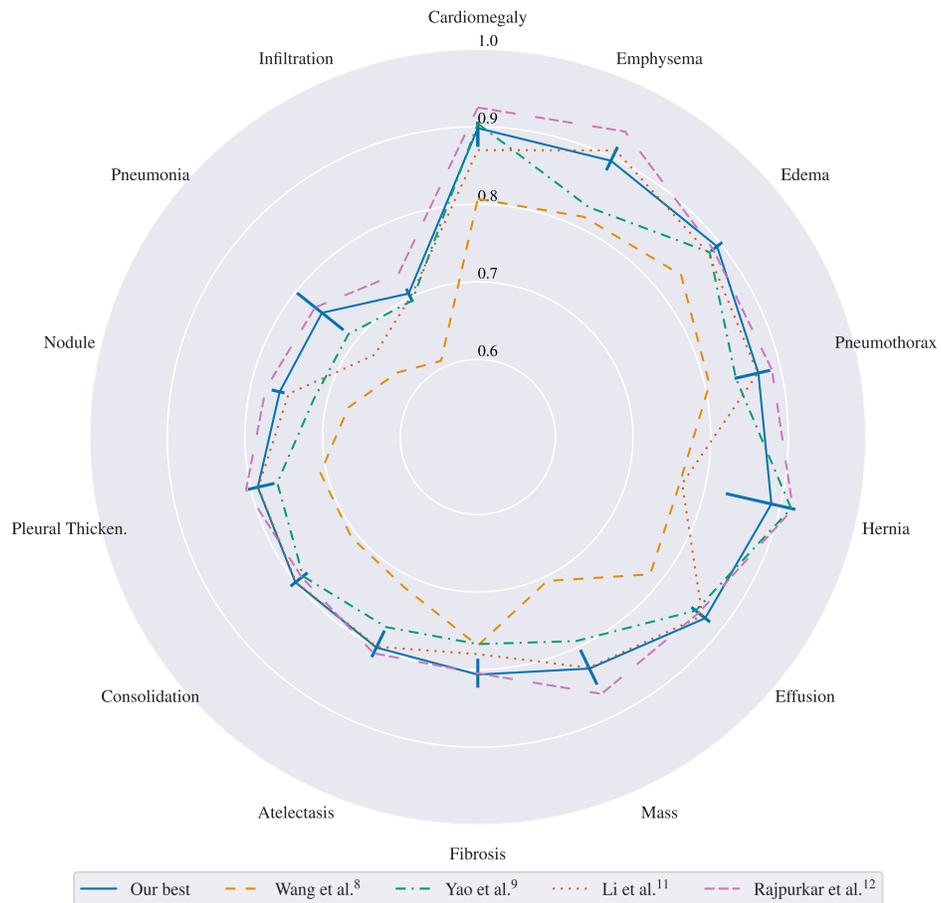}
		
		\caption{Comparison of our best model to other groups. We sort the pathologies with increasing average AUC over all groups. For our model, we report the \rchange{min}{minimum} and \rchange{max}{maximum} over all folds as error bar to illustrate the effect of splitting.}
		\label{fig:compare}
		\vspace{-1.5em}
	\end{figure}
}	
	\rmove{
	\begin{table}
		\setlength{\tabcolsep}{5pt}
		\centering
		\caption[Comp]{\rtag{R1.5-6}\rtag{R2.5-7}\rnew{AUC result overview for our experiments on the official split. In this table, we present results for our best performing architecture with different depth (i.e. ResNet38-large-meta, ResNet50-large-meta, ResNet101-large-meta) and compare them to other groups. Additionally we provide an average AUC over all pathologies in the last row. Bold text emphasizes the overall highest AUC value.}}
		\label{tab:compareOfficial}
		\begin{tabular}{l c c c | c c c }
			\noalign{\smallskip}
			& & & & \multicolumn{3}{c}{``-large-meta''} \\
			\textbf{Pathology} & Wang et al.\cite{Wang2017} & Yao et al.\cite{yao2018weakly} &  Guendel et al.\cite{guendel2018learning} & ResNet-38 & ResNet-50 & ResNet-101 \\
			\noalign{\smallskip}
			\hline
			\noalign{\smallskip}	
			\textbf{Cardiomegaly}
			& $0.810$
			& $0.856
			$
			& $\textbf{0.883}
			$
			& $0.875
			$
			& $0.877
			$
			& $0.865
			$
			\\
			
			\textbf{Emphysema}
			& $0.833
			$
			& $0.842
			$
			& $\textbf{0.895}
			$
			& $\textbf{0.895}
			$
			& $0.875
			$
			& $0.868
			$
			\\
			
			\textbf{Edema}
			& $0.805
			$
			& $0.806
			$
			& $0.835
			$
			& $\textbf{0.846}
			$
			& $0.842
			$
			& $0.828
			$
			\\
			
			\textbf{Hernia}
			& $0.872
			$
			& $0.775
			$
			& $0.896
			$
			& $\textbf{0.937}
			$
			& $0.916
			$
			& $0.855
			$
			\\
			
			\textbf{Pneumothorax}
			& $0.799
			$
			& $0.805
			$
			& $\textbf{0.846}
			$
			& $0.840
			$
			& $0.819
			$
			& $0.839
			$
			\\
			
			\textbf{Effusion}
			& $0.759
			$
			& $0.806
			$
			& $\textbf{0.828}
			$
			& $0.822
			$
			& $0.818
			$
			& $0.818
			$
			\\
			
			\textbf{Mass}
			& $0.693
			$
			& $0.777
			$
			& $\textbf{0.821}
			$
			& $0.820
			$
			& $0.810
			$
			& $0.796
			$
			\\
			
			\textbf{Fibrosis}
			& $0.786
			$
			& $0.743
			$
			& $\textbf{0.818}
			$
			& $0.816
			$
			& $0.800
			$
			& $0.778
			$
			\\
			
			\textbf{Atelectasis}
			& $0.700
			$
			& $0.733
			$
			& $\textbf{0.767}
			$
			& $0.763
			$
			& $0.755
			$
			& $0.747
			$
			\\
			
			\textbf{Consolidation}
			& $0.703
			$
			& $0.711
			$
			& $0.745
			$
			& $\textbf{0.749}
			$
			& $0.742
			$
			& $0.734
			$
			\\
			
			\textbf{Pleural Thicken.}
			& $0.684
			$
			& $0.724
			$
			& $0.761
			$
			& $\textbf{0.763}
			$
			& $0.742
			$
			& $0.739
			$
			\\
			
			\textbf{Nodule}
			& $0.669
			$
			& $0.724
			$
			& $\textbf{0.758}
			$
			& $0.747
			$
			& $0.736
			$
			& $0.738
			$
			\\
			
			\textbf{Pneumonia}
			& $0.658
			$
			& $0.684
			$
			& $\textbf{0.731}
			$
			& $0.714
			$
			& $0.703
			$
			& $0.694
			$
			\\
			
			\textbf{Infiltration}
			& $0.661
			$
			& $0.673
			$
			& $\textbf{0.709}
			$
			& $0.694
			$
			& $0.694
			$
			& $0.686
			$
			\\
			
			\hline
			\noalign{\smallskip}
			
			\textbf{Average}
			& $0.745$
			& $0.761$
			& $\textbf{0.807}$
			& $0.806$
			& $0.795$
			& $0.785$
			\\
			
			\textbf{No Findings}
			& -
			& -
			& -
			& $0.727$
			& $0.725$
			& $0.720$
			\\
			
		\end{tabular}
	\end{table}
}
	
	\newpage
	
	\section{Discussion and Conclusion}
	\label{sec:discuss}
	We present a systematic evaluation of different approaches for CNN-based X-ray classification on ChestX-ray14. While satisfactory results are obtained with networks optimized on the ImageNet dataset, the best overall results can be reported for the model that is exclusively trained with CXRs and incorporates non-image data (i.e. view position, patient age, and gender).
	
	Our optimized \rchange{ResNet-50}{ResNet-38-large-meta} architecture achieves state-of-the art results in \rchange{four}{five} out of fourteen classes compared to \rchange{Rajpurkar et al.\cite{Rajpurkar2017}}{Guendel et al.\cite{guendel2018learning}} (who had state-of-the-art results in all fourteen classes\rchange{). At the same time, a substantial variability in the results can be observed when different splits are considered. This becomes especially apparent for ``Hernia'', the class with the fewest samples in the dataset (see also Figure~\ref{fig:compare}}{on the official split)}. For other classes even higher scores are reported in the literature (see e.g. Rajpurkar et al.\cite{Rajpurkar2017}). However, a comparison of the different CNN methods with respect to their performance is inherently difficult, as most evaluations have been performed on individual (random) partitions of the datasets. \rnew{We observed substantial variability in the results when different splits are considered. This becomes especially apparent for ?Hernia?, the class with the fewest samples in the dataset (see also Figure~\ref{fig:compare}).}
	
	While the obtained results suggests that the training of deep neural networks in the medical domain is a viable option as more and more public datasets become available, the practical use of deep learning in clinical practice is still an open issue.
	In particular for the ChestX-ray14 datasets, the rather high label noise\cite{Wang2017} of 10\% makes an assessment of the true network performance difficult. Therefore, a clean testset without label noise is needed for clinical impact evaluation. As discussed by Oakden-Rayner \cite{Luke2017}, the quality of the (automatically generated) labels and their precise medical interpretation may be a limiting factor addition to the presence of treated findings. Our Grad-CAM results proves Oakden-\rchange{Rayner}{Rayner's} concerns about the ``pneumothorax'' label. In a clinical setting, i.e. for the detection of critical findings, the focus would be on the reliably identification of acute cases of pneumothorax, while a network trained on ChestX-ray14 would also respond to cases with a chest drain. 
	
	Future work will include investigation of other model architectures, new architectures for leveraging label dependencies and incorporating segmentation information.

	\bibliographystyle{splncs03}
	\bibliography{references}
	
\end{document}